\documentclass[10pt,twocolumn,letterpaper]{article}

\usepackage{iccv}              

\definecolor{iccvblue}{rgb}{0.21,0.49,0.74}
\usepackage[pagebackref,breaklinks,colorlinks,allcolors=iccvblue]{hyperref}

\usepackage{hyperref}       
\usepackage{url}            
\usepackage{booktabs}       
\usepackage{amsfonts}       
\usepackage{nicefrac}       
\usepackage{microtype}      

\usepackage{xcolor}
\usepackage{colortbl}
\usepackage{amsmath}

\definecolor{Gray}{gray}{0.85}
\definecolor{LightCyan}{rgb}{0.88,1,1}
\newcolumntype{a}{>{\columncolor{Gray}}c}

\usepackage{multirow}
\usepackage{adjustbox}
\usepackage{array}
\usepackage{wrapfig,lipsum}

\usepackage{pifont}
\usepackage{graphicx}
\usepackage{arydshln}
\usepackage{tikz}

\definecolor{darkgreen}{rgb}{0.2, 0.7, 0.1}

\definecolor{Gray}{gray}{0.8}
\definecolor{LG}{gray}{.92}

\newcommand*\colourcheck[1]{%
  \expandafter\newcommand\csname #1check\endcsname{\textcolor{#1}{\ding{51}}}%
}
\colourcheck{green}
\colourcheck{darkgreen}

\usepackage{multirow}
\usepackage{adjustbox}
\usepackage{array}
\usepackage{wrapfig,lipsum,booktabs}
\usepackage{amsmath,amssymb}
\usepackage{enumitem}
\definecolor{darkgreen}{rgb}{0.2, 0.7, 0.1}
\newcommand*\colourxmark[1]{%
  \expandafter\newcommand\csname #1xmark\endcsname{\textcolor{#1}{\ding{55}}}%
}
\colourxmark{red}
\usepackage{arydshln}

\usepackage{stmaryrd}
\usepackage{trimclip}

\makeatletter
\DeclareRobustCommand{\shortto}{%
  \mathrel{\mathpalette\short@to\relax}%
}

\newcommand{\short@to}[2]{%
  \clipbox{{.3\width} 0 0 0}{$\m@th#1\vphantom{+}{\shortrightarrow}$}%
  }
\makeatother
\usepackage{makecell}

\usepackage{adjustbox}
\usepackage{graphicx}
\usepackage{fix-cm}
\makeatletter
\def\thickhline{%
  \noalign{\ifnum0=`}\fi\hrule \@height \thickarrayrulewidth \futurelet
   \reserved@a\@xthickhline}
\def\@xthickhline{\ifx\reserved@a\thickhline
               \vskip\doublerulesep
               \vskip-\thickarrayrulewidth
             \fi
      \ifnum0=`{\fi}}
\makeatother
\newlength{\thickarrayrulewidth}
\usepackage{setspace}
\usepackage{bm}
\usepackage{tikz}

\usepackage{enumitem}

\def\paperID{5403} 
\def\confName{ICCV}
\def\confYear{2025}

\title{From Sharp to Blur: Unsupervised Domain Adaptation for 2D Human Pose
Estimation Under Extreme Motion Blur Using Event Cameras}

\author{Youngho Kim$^{*}$, Hoonhee Cho$^{*}$, and Kuk-Jin Yoon  \\
KAIST\\
{\tt\small \{kmax2001,gnsgnsgml,kjyoon\}@kaist.ac.kr}
}

\begin{document}
\maketitle
\begin{abstract}
Human pose estimation is critical for applications such as rehabilitation, sports analytics, and AR/VR systems. However, rapid motion and low-light conditions often introduce motion blur, significantly degrading pose estimation due to the domain gap between sharp and blurred images. Most datasets assume stable conditions, making models trained on sharp images struggle in blurred environments. To address this, we introduce a novel domain adaptation approach that leverages event cameras, which capture high temporal resolution motion data and are inherently robust to motion blur. Using event-based augmentation, we generate motion-aware blurred images, effectively bridging the domain gap between sharp and blurred domains without requiring paired annotations. Additionally, we develop a student-teacher framework that iteratively refines pseudo-labels, leveraging mutual uncertainty masking to eliminate incorrect labels and enable more effective learning. Experimental results demonstrate that our approach outperforms conventional domain-adaptive human pose estimation methods, achieving robust pose estimation under motion blur without requiring annotations in the target domain. Our findings highlight the potential of event cameras as a scalable and effective solution for domain adaptation in real-world motion blur environments. Our project codes are available at \url{https://github.com/kmax2001/EvSharp2Blur}.
\end{abstract}
    
\section{Introduction}
\label{sec:intro}

\def\thefootnote{*}\footnotetext{The first two authors contributed equally.}\def\thefootnote{\arabic{footnote}} 

Human pose estimation identifies key body joints or limbs, essential for applications in rehabilitation, sports analytics, and AR/VR systems. Despite the dynamic nature of human motion, most datasets assume stable conditions, providing only clean, well-conditioned data. As a result, models trained on such datasets struggle with motion blur due to the significant domain gap between sharp and blurred images. Conventional cameras rarely capture paired sharp-blurred images, making domain adaptation challenging. While dual-camera setups with beam splitters~\cite{cho2025benchmark, duan2023eventaid} offer precise annotations, they require expertise and are impractical for general user use. Addressing motion blur without such complex systems remains a key challenge in pose estimation.

\begin{figure}[t]
    \centering
    \includegraphics[width=0.92\linewidth]{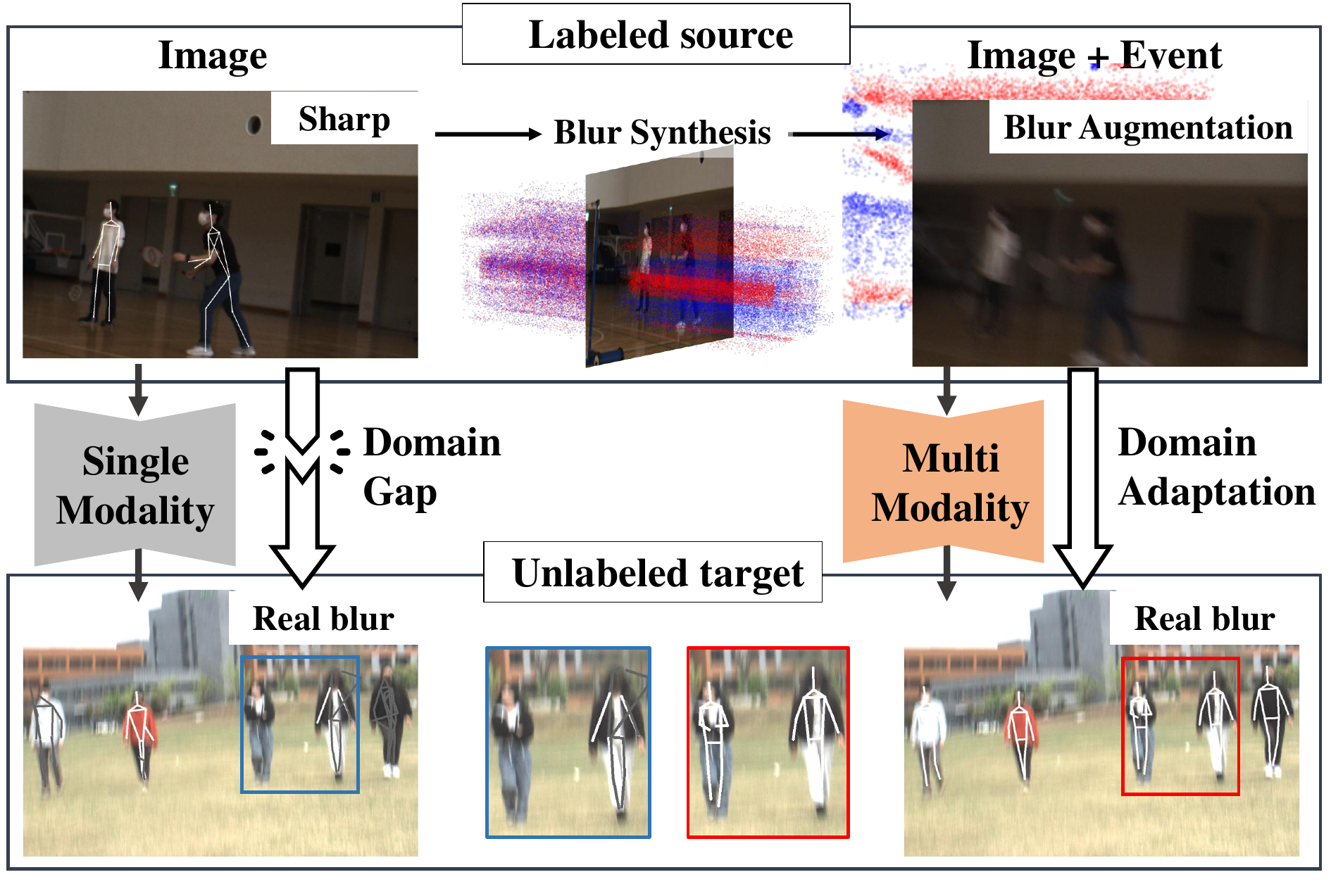}
    \vspace{-8pt}
    \caption{Image-based networks trained on sharp images face performance degradation in blurred images due to the domain gap. In contrast, our method uses event data for blur augmentation and achieves superior performance in the blurred domain through multi-modality adaptation.}
    \label{fig:teaser}
    \vspace{-10pt}
\end{figure}

Unlike conventional cameras, event cameras~\cite{gallego2020event,yang2015dynamic} operate at a high frame rate, making them resistant to motion blur. In contrast to standard cameras, which capture images at fixed intervals, event cameras detect per-pixel intensity changes, producing sparse but temporally rich data. Previous research has utilized event cameras for motion deblurring~\cite{sun2024motion, yang2024motion, Huang2023EventbasedSL, sun2022event, cho2023non}, image reconstruction~\cite{cannici2024mitigating, wang2024revisit, weng2021event, yang2023learning}, and perceptions under low-light conditions~\cite{xia2023cmda, li2024event, yao2024event, jeong2024towards, liang2023coherent, liu2024seeing, liang2024towards, kim2024towards}. 
However, efforts to bridge the motion blur domain gap, particularly in multi-person pose estimation, remain unexplored.

In this study, we focus on transferring knowledge from a sharp domain with ground-truth annotations to a motion-blurred environment where pose annotations are difficult to obtain. Unsupervised domain adaptation relies on generating reliable pseudo-labels~\cite{xia2023cmda, zheng2024eventdance, cho2024tta} in the target domain. However, the large domain gap between sharp and blurred images makes this a challenging task, and traditional cameras alone lead to suboptimal adaptation. To address this, we introduce a novel method for adapting 2D multi-human pose estimation from a sharp domain to a blurred domain using event cameras, which capture continuous motion data over time. Unlike static sharp images that lack motion information, event cameras provide detailed motion data, enabling each time slice to capture the full pixel movement within that period. As illustrated in Fig.~\ref{fig:teaser}, we utilize this characteristic to generate motion-aware blurred images through event-based augmentation, using event cameras to bridge the gap between the two domains. By employing these synthesized motion-aware blurred images, we effectively reduce the domain discrepancy, allowing the model to maintain reasonable performance in the target domain even when trained exclusively on the source domain.

In addition, we focus on the complementary strengths of images and event data. Images capture dense spatial information, effectively preserving semantic details in sharp domains and often helping in blurred domains as well. In contrast, event cameras capture sparse motion cues along moving edges, enabling robust detection even under extreme motion but often suffering from false detections due to their lack of spatial density. 
To leverage the strengths of both modalities, we design a teacher network with a sub-network that takes different types of modalities as input and a refinement module that adaptively utilizes the strengths of each output of sub-networks. This module enables scene-specific pose estimation by balancing the spatial richness of images and the motion-invariant edge features of events, ultimately generating high-quality pseudo-labels for adaptation.

Additionally, we propose a method that jointly employs a student-teacher framework for pseudo-label generation. We alternately adapt the student and teacher networks to the target domain, with the teacher network leveraging the student network to generate more reliable pseudo-labels for adaptation. In turn, the enhanced teacher network further trains the student network, creating a feedback loop that improves overall performance.

Our approach, the first to leverage event cameras to bridge the domain gap caused by motion blur, enables effective adaptation from the sharp domain to the blur domain without ground-truth annotations in the blur domain. During the adaptation process, our method minimizes performance degradation in the source domain, ultimately achieving superior results in both the sharp and blur domains.

\section{Related Works}
\label{sec:related_work}
\noindent
\textbf{Domain-adaptive Human Pose Estimation.} 
There are two paradigms in human pose estimation: bottom-up and top-down approaches. The top-down method detects the human first, and then estimates keypoint locations~\cite{chen2018cascaded, sun2019deep, tian2019directpose, papandreou2017towards, fang2017rmpe, wei2016convolutional, newell2016stacked}, while the bottom-up method detects keypoints first, and then groups them into poses~\cite{cao2019openpose, cheng2020higherhrnet, newell2017associative, sun2020bottom, pishchulin2016deepcut, insafutdinov2016deepercut, iqbal2016multi, kreiss2019pifpaf}

Training 2D human pose estimation models requires large labeled datasets, which are labor-intensive and time-consuming. To address this, synthetic datasets are often used~\cite{kim2022unified, martinez2018investigating}, but a domain gap remains, limiting generalization across datasets~\cite{ganin2015unsupervised, ragab2022self}. To bridge this gap, data augmentation techniques are commonly applied to mitigate these gaps~\cite{kim2022unified, chen2024domain, martinez2018investigating, ai2024domain, crescitelli2020poison}.
Despite advances in domain adaptation for human pose estimation, little has been explored in the challenging blurry domains. Estimation of pose in blurred images~\cite{zhao2023human} is difficult due to the complexities of the augmentation needed to bridge the domain gap. Specifically, generating blur based on motion from a single image requires sophisticated methods that account for object motion, which is particularly challenging. To this end, we propose a novel event-based augmentation method that considers actual motion to address these challenges.

\noindent
\textbf{Pseudo-label Refinement.} 
To train on an unlabeled dataset, teacher-student frameworks are often adopted, where teacher networks generate pseudo labels to guide student network training~\cite{lee2013pseudo, berthelot2019mixmatch, wu2024leod, xia2023cmda, huang2023semi, xie2021empirical, zhuang2022semi, cho2024finding, fang2018weakly, cho2023label}. To enhance the reliability of pseudo labels, various refinement techniques has been explored. For example, MixMatch~\cite{berthelot2019mixmatch} applies data augmentation at various levels and aggregates predictions to improve robustness. LEOD~\cite{wu2024leod} employs a high confidential threshold to obtain more reliable pseudo labels and utilizes tracking-based post processing to filter out temporally inconsistent and noisy labels. SSPCM~\cite{huang2023semi} used pseudo-label correction module that selects only consistent prediction from dual networks. We also propose a learning-based pseudo-label refinement module that leverages outputs from multiple models, each trained on a different modality.

\begin{figure*}[t!]
\begin{center}
\includegraphics[width=.88\linewidth]{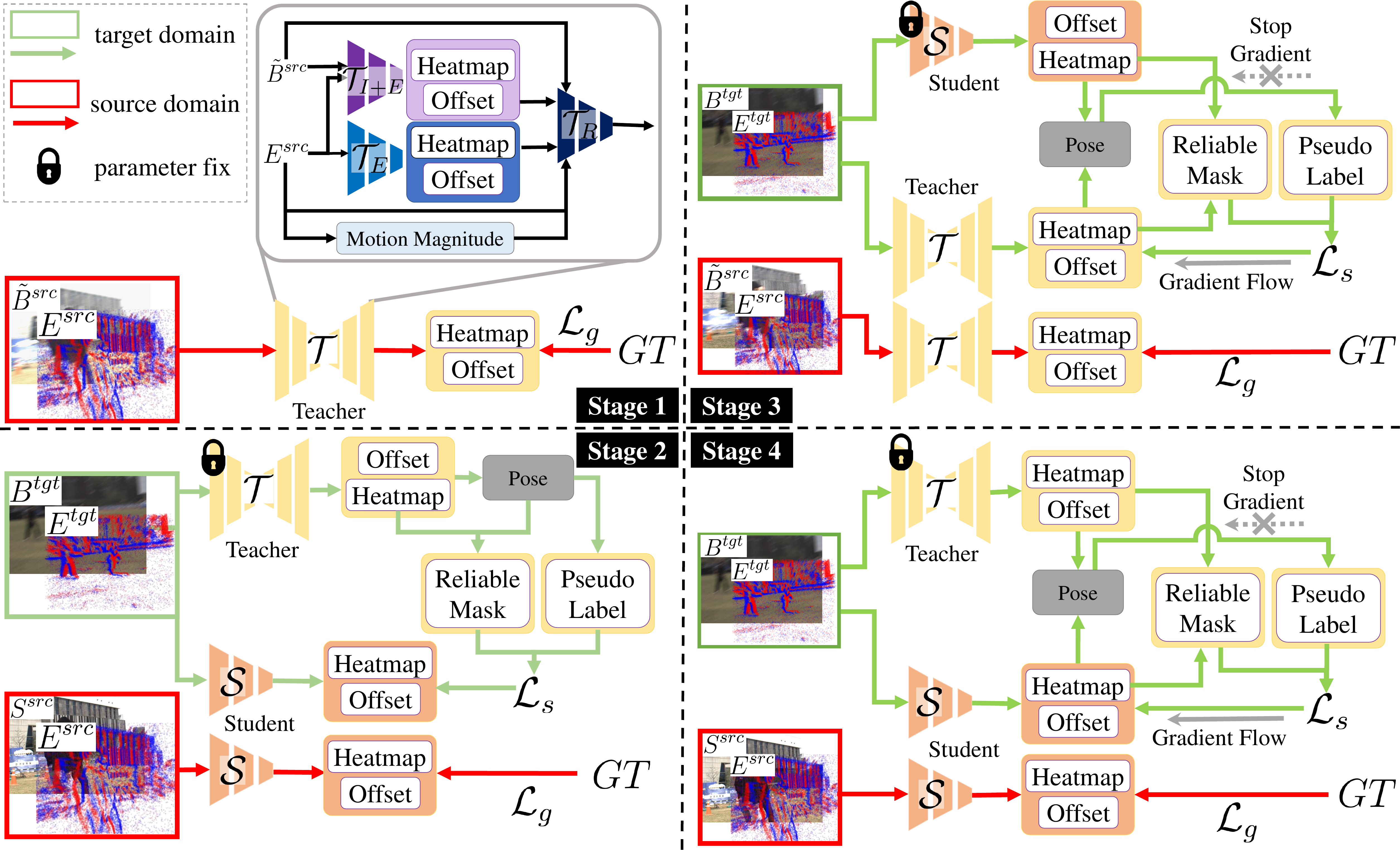}
\vspace{-5pt}
\caption{Overall framework of the proposed domain-adaptive human pose estimation. The proposed method consists of four sequential stages. Stage 1: Teacher Network Pre-training, Stage 2: Student Network Training with Pseudo-Labels, and Stages 3 and 4: Mutual Uncertainty Masking for the Teacher and Student Networks. $GT$ denotes the groud-truth human pose.
}

\label{fig:overall_framework}
\end{center}
\vspace{-19pt}
\end{figure*}

\noindent
\textbf{Event-based Human Pose Estimation.}
Event cameras are robust to motion blur and have low latency, which has led to growing attempts to apply them to human pose estimation. Specifically, there have been efforts to estimate dynamic features such as hand pose~\cite{jiang2024evhandpose, jiang2024complementing, rudnev2021eventhands, nehvi2021differentiable}, leveraging the ability of event cameras to perform well in extreme lighting and blur conditions. Additionally, datasets and approaches for whole-body pose estimation~\cite{zou2021eventhpe, calabrese2019dhp19, xu2020eventcap, yan2024reli11d, millerdurai2025eventego3d++, millerdurai2024eventego3d} are emerging. Recently, the first RGB and event dataset~\cite{cho2025benchmark}, including pose annotations for multi-human poses, has emerged, encompassing extreme blur and low-light conditions. Building on these recent efforts using event cameras to tackle the blur domain, we are the first to attempt bridging the gap between sharp and blur domains in multi-human pose estimation.

\noindent
\textbf{Event-based Domain Adaptation.} 
Event cameras have varying distributions depending on the illumination and the device, and approaches for adapting models to these changing distributions~\cite{jian2023unsupervised, planamente2021da4event, cho2024tta, kim2022ev, hu2020learning} have been continually proposed. Additionally, there have been efforts~\cite{wang2021evdistill, cho2023learning, sun2022ess, messikommer2022bridging, zheng2024eventdance, cannici2021n} to adapt from the image domain to the event domain to compensate for the lack of sensor data, as well as approaches to adapt perception models for low-light conditions~\cite{xia2023cmda}. As another new challenge, we use event cameras as a bridge between motion blur and sharp domains, and for the first time, we attempt to adapt to the blur domain using annotations only from the sharp domain.

\section{Methods}
\label{sec:methods}

\subsection{Problem Definition and Overall Framework}

Domain adaptive 2D human pose estimation aims to achieve high performance on a target domain with a different data distribution, using only labeled source domain data. We assume the source domain consists of sharp images, $S^{src}$, with annotations, while the target domain contains blur images, $B^{tgt}$, without annotations. 

Event data is present in both domains, with \( E^{src} \) representing events from the source domain and \( E^{tgt} \) from the target domain. Each image corresponds to a single frame, and our goal is to address multi-human 2D pose estimation.

Our network structure follows DEKR \cite{geng2021dekr}. Each network simultaneously outputs a center heatmap $\mathcal{H_C}$, offset $\mathcal{O}$, and keypoint heatmaps $\mathcal{H} _k$, where $k$ represents the index of each keypoint. The center heatmap detects the human's center whose location is the local argmax of the center heatmap. Offsets, defined as $\mathcal{O} = \{{p_c^{i} - p_{1}^{i}, p_c^{i} -p_{2}^{i},\dots, p_c^{i} - p^{i}_{k}, \dots ,p_c^{i} - p^{i}_{K}  }\}$, are displacement of 
each the human's center $p_c^i$  
from keypoints location ${p^{i}_{k}}$ 
where ${p^{i}_{k}} \in {R^2}$ represents the 2D location of the $k$-th keypoints for the $i$-th person. The keypoint heatmap is used for scoring and ranking the regressed poses by averaging the heat value on each keypoint location.
Following \cite{geng2021dekr}, supervised loss for heatmap and offset is defined as 
\begin{equation}    
\mathcal{L}_g = L_{\mathcal{H}} + \lambda_g \cdot L_{\mathcal{O}}
\label{equ:sup_loss}
\end{equation} 
$L_{\mathcal{H}}$ means a MSE loss of the predicted heatmap and $L_{\mathcal{O}}$ means a smooth L1 loss of the offset map. $\lambda_g$ is a hyper-parameter and set as 0.03 in our experiment.

As shown in Fig.~\ref{fig:overall_framework}, our overall training process is divided into four main stages. \textbf{(1) Teacher Network Pretraining:} We train a high-capacity teacher network that leverages multi-modal learning. To ensure superior performance on the target domain using only source domain data, we propose event-based augmentation techniques. \textbf{(2) Student Network Training with Pseudo-Labels:} Using the pretrained teacher network, we generate pseudo-labels to train the student network $\mathcal{S}$. The goal is to enable $\mathcal{S}$ to learn effectively from both the source and target domains. \textbf{(3) Mutual Uncertainty Masking for Teacher Network:} The trained student and teacher networks are then used together to mutually mask uncertain predictions, generating more reliable pseudo-labels. \textbf{(4) Mutual Uncertainty Masking for Student Network:} The enhanced teacher is used again to generate pseudo-labels with student network, further refining the student network’s performance.

\subsection{Teacher Network Pretraining}

In the target domain, where labels are unavailable, recent studies~\cite{ai2024domain, chen2024domain} have used teacher networks to generate pseudo-labels for effective training on unlabeled data. A higher-capacity of teacher network and robustness to domain shifts generate better pseudo-labels. We propose a robust teacher network trained with motion-aware blur augmentation using event data to ensure strong performance in the target domain, even when trained solely on the source domain. Additionally, the teacher network incorporates multiple sub-networks that use various modalities, with a refinement network.

\begin{figure}[t]
    \centering
    \includegraphics[width=0.94\linewidth]{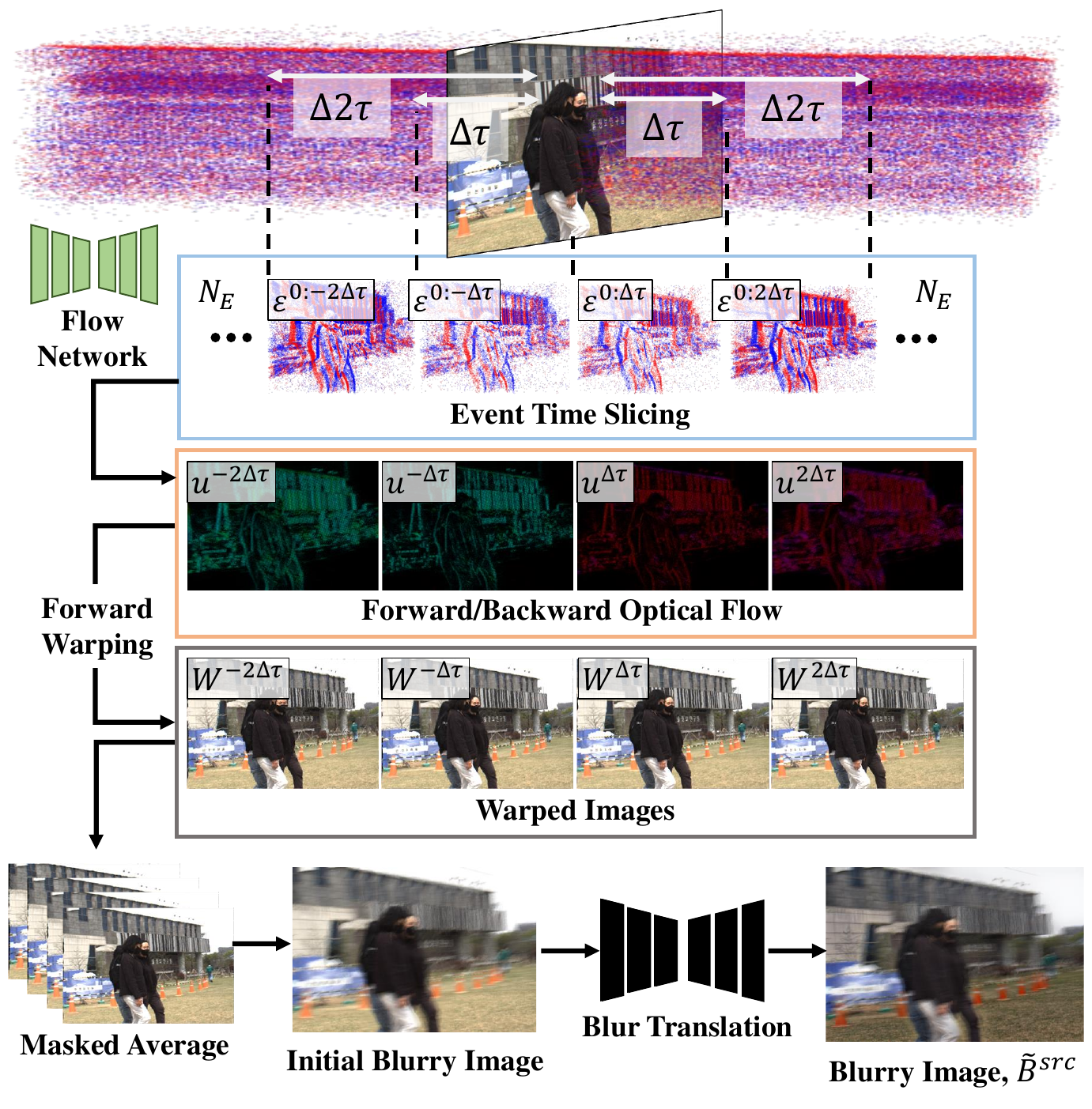}
    \vspace{-10pt}
    \caption{Motion-aware event-based blur augmentation.}
    \label{fig:motion_aug}
    \vspace{-10pt}
\end{figure}

\noindent
\textbf{Motion-aware Event-based Blur Augmentation.}
Our setup deals with the scenario of a single image, similar to the domain-adaptive multi-human pose estimation~\cite{ai2024domain, kim2022unified}. 

Generating blur from a single image is extremely challenging. As an alternative, one can randomly generate a blur kernel and apply it to the image to simulate blur. However, this approach does not follow the actual principles of motion blur. Since it is not generated based on any motion, it is particularly ineffective in representing blur caused by object movement.
To solve this problem, we propose the blur augmentation method that considers the actual motion by leveraging event data. Real-world motion blur images~\cite{rim2020real, duan2023eventaid, kim2024frequency, kim2024cross} are formed by the continuous accumulation of pixel information during the exposure time. 

To replicate this, we compute pixel movement from events and shift pixels accordingly. For this, we train an event-based optical flow network with a self-supervised loss~\cite{gallego2018unifying, paredes2020back, cho2024temporal}.

As shown in Fig.~\ref{fig:motion_aug}, we estimate motion by slicing the event data during a time duration longer than the exposure time of the sharp image, with the center timestamp of the sharp image serving as a reference.

We divide the event slice set into \( 2N_E \) segments over \( \Delta\tau \) and estimate the flow for each slice:
$
\left\{ u^{(t\Delta\tau)} \mid t = -N_E, \dots, N_E \right\},
$
based on the corresponding event sequences:
$
\left\{ \varepsilon^{(0:t\Delta\tau)} \mid t = -N_E, \dots, N_E \right\}.
$
The estimated flows are then used for forward warping on a sharp reference image to generate the warped images:
$
\left\{ W^{(t\Delta\tau)} \mid t = -N_E, \dots, N_E \right\}.
$
Forward warping often introduces holes, so instead of averaging across all images, we handle these gaps as follows:
\begin{equation}
\tilde{B}^{src}(x,y) =  \frac{\sum _{t} W_{t} (x,y) \cdot \delta_{t}(x,y)}{\sum \delta_t(x,y) + \epsilon}
\label{equ:blur_avg}
\end{equation}
where \( W_{t}(x,y) \) denotes the intensity of the \( t \)-th warped image at location \( (x,y) \). If the \( t \)-th sharp image contains a hole at \( (x,y) \), then \( \delta _ {t}(x,y) = 0 \); otherwise, \( \delta _ {t}(x,y) = 1 \). The small constant \( \epsilon \) prevents division by zero.

This process still generates a discrete blur image. To make it more realistic, we apply a blur translation loss~\cite{pham2024blur2blur} to the generated blur in order to reduce the distribution gap between discrete and continuous blurry images.

\noindent
\textbf{Multi-modal teacher network.}

Event and image data have distinct characteristics. Event-based networks are robust to motion blur, maintaining performance across domains. In contrast, image-based networks are highly sensitive to motion blur but capture rich intensity information. When trained with annotated blur images and combined with event data, network using both image and event outperforms purely image- and event-based approaches.

To leverage the advantages of each modality, we design the overall teacher network, $\mathcal{T}_{\mathcal{M}}$, as a structure composed of sub-teacher networks, each based on different modalities. They are constructed based on~\cite{geng2021dekr, cho2025benchmark}, where $\mathcal{M}$ $\in$ $\{I+E, E\}$ means the modality of the input of the network. As shown in Stage 1 of Fig.~\ref{fig:overall_framework}, our teacher network consists of two sub-networks: an only event-based network, $\mathcal{T}_E$, and an image-event fusion network, $\mathcal{T}_{I+E}$. To combine the results of these two networks, we use an additional refinement network, $\mathcal{T}_R$. The refinement network generates the final heatmap and offset using image and event inputs, along with those from each sub-network. Since the teacher network is used only during training, we enhance results by providing additional guidance. Specifically, the refinement network receives the magnitude of optical flow, computed from events. The refinement network, $\mathcal{T}_R$, concatenates all inputs along the channel dimension and fuses them through a channel attention mechanism~\cite{zamir2022restormer, sun2022event}.

To ensure effective adaptation of the teacher network to the blur domain, Stage 1 excludes sharp images. Instead, the teacher network is trained with synthesized blur images generated through augmentation, using source domain annotations and supervised by \( \mathcal{L}_g \) (Eq.~(\ref{equ:sup_loss})) during this process.

\subsection{Student Network Training with Pseudo-Labels}
\label{sec:stage2}
In Stage 2, the goal is to train the student network using the accurate annotations from the source domain and pseudo-labels from teacher network in the target domain.
The student network takes image and event data as inputs and has the same architecture as \( \mathcal{T}_{I+E} \).
Using the teacher network trained in Stage 1, which is partially adapted to the blur domain through augmentation and takes advantages of multi-modal data through refinement, we generate pseudo-labels for the target dataset.

Firstly, we estimate center location from center heatmap, $\mathcal{H}_c ^{\mathcal{T}}$ and generate each pose $P_i^{\mathcal{T}}$ = $\{p^{i}_k , p^{i}_c\} _{k \in \{1,.., K\}}$ using offset, $\mathcal{O}_k^{\mathcal{T}}$ of each keypoint. $p^{i}_k$ means keypoint location of $k$-th keypoint of $i$-th pose. Poses, $\{ P_i\}$, which is set of each pose  are then filtered out based on non-maximum suppression (NMS)~\cite{hosang2017learning}. 
However, the performance can degrade significantly when noisy or incorrect labels are provided during training. To avoid this, we introduce pixel-wise masking, $M$, to calculate the loss function. By averaging keypoint heatmap values on each keypoint, we calculate confidence score of each pseudo pose as follows:
\begin{equation}
    C(P_i) = \mathcal{H}^\mathcal{T}_c(p^i_c) * \sum^{K}_{k=1} {\mathcal{H}^{\mathcal{T}}_k(p^i_k)} /K
    \label{equ:confidence}
\end{equation}

Masking is applied when the confidence score is below \( th \), and the confidence-based heatmap mask $M^\mathcal{H}$ is defined as:

\begin{equation}
d_k{(x,y)}^\mathcal{H} := \{i | (x,y) \in near(p^i_k) \} \\
\end{equation}

\begin{equation}
M_{k}(x, y) ^{\mathcal{H}} = \prod_{i \in d_k(x,y)^{\mathcal{H}}} m, \ 
m =
\begin{cases} 
  1, &  \ C(P_i)\geq th \\ 
  0, & else
\end{cases}
\label{equ: one_teacher_masking}
\end{equation}

If $d_k(x,y) ^{\mathcal{H}}= \emptyset$, the pixel $(x,y)$ is considered as a background and $M_k(x,y) ^{\mathcal{H}}$ is set to 0.1, following previous work~\cite{geng2021dekr}. 
The mask $M_k(x,y)^{\mathcal{H}}$ filters out where unreliable pseudo pose lies not to calculate loss of the keypoint $k$ at $(x,y)$. This helps to exclude unreliable pixels from both the heatmap and offset, ensuring stable training in the target domain.
$M_{k}(x, y) ^{\mathcal{O}}$ is defined below.
\begin{equation}
    M_{k}(x, y) ^{\mathcal{O}} = { 1 \over {\min (Z_{i})}}
\end{equation}
$Z_i$ = $\{\sqrt{H^2_i+W^2_i } | (x,y) \in near(p_c^i) \}$ is the set of the size of the corresponding person instance, and $H_i$ and $W_i$ are the height and width of the instance box \cite{geng2021dekr}. If $Z_i = \emptyset$, $M_{k}(x, y) ^{\mathcal{O}} = 0$.
To compute the loss during student network training with the mask applied, where $\mathcal{O}$, $\mathcal{H}$ are offset prediction and heatmap prediction of student network, we use the following approach:
\begin{equation}
L_{\mathcal{H}} = M^{\mathcal{H}} *|| \mathcal{H} - \mathcal{H}^{\mathcal{T}} || ^2 _2 
\label{equ: masked_heatmap_loss}
\end{equation}
\begin{equation}
    L_{\mathcal{O}} =  M^{\mathcal{O}} * \text{Smooth}_{L_1}(\mathcal{O} - \mathcal{O} _{k}^{\mathcal{T}})
\label{equ: masked_offset_loss}
\end{equation}

\subsection{Mutual Uncertainty Masking}
\noindent
\textbf{Teacher Network Training.}
Through Stages 1 and 2, we obtain a student network partially adapted to the target domain by using pseudo-labels generated by teacher network. These pseudo-labels depend on the performance of the teacher network so using a higher-performing teacher network can generate higher-quality pseudo-labels. However, since the pre-trained teacher network has not seen real blur yet in the target domain, there is still room for performance improvement through adaptation to the target domain. 

In the blur domain, noise often causes false detections and missed detections, making it unreliable to fully trust a single network. Therefore, we propose mutual uncertainty masking, where the confidence scores from both the teacher and student networks are considered to mask out unreliable regions. Specifically, we first generate teacher poses, \( \{P^\mathcal{T}_i\} \), from the teacher network and student poses, \( \{P^{\mathcal{S}}_i \} \), from the student network. In Stages 3 and 4, we compute another confidence score, \( C'(P_i) \), for the poses of each network based on the heatmap of the other network. The equation below calculates the confidence scores using the student network, $\mathcal{S}$:
\begin{equation}
        C'(P^{\mathcal{S}}_i) = \mathcal{H} ^{\mathcal{T}}_c(p^i_c) * \sum^{K}_{k=1} {\mathcal{H}_k^{\mathcal{T}}(p^i_k)} /K
        \label{equ:confidence_crosscheck}
\end{equation}
\begin{equation}
    C(P^{\mathcal{S}}_i) = \mathcal{H} ^{\mathcal{S}}_c(p^i_c) * \sum^{K}_{k=1} {\mathcal{H}_k^{\mathcal{S}}(p^i_k)} /K
        \label{equ:confidence}
\end{equation}
For scores of poses by $\mathcal{T}$, $C(P_i^{\mathcal{T}})$ should be calculated
with $\mathcal{H}_c^{\mathcal{T}}$, $\mathcal{H}_k^{\mathcal{T}}$ and $C'(P^{\mathcal{T}})$ should be calculated with $\mathcal{H}_c^{\mathcal{S}}$, $\mathcal{H}_k^{\mathcal{S}}$.

\begin{table*}[!t]
    \setlength\tabcolsep{4.0pt}
    \centering
    \caption{Multi-person pose estimation evaluated on the EHPT-XC dataset. 
    To showcase the robustness of method beyond its performance in the target domain, we also report results in the source domain. The methods in top six rows represent oracle networks that leverage annotations from the target domain. `$\text{-}$' indicates that evaluation was not performed, as the teacher network did not utilize sharp images of the source domain for training. Bold and underline indicate the best and second-best performances among networks trained only on the source domain, respectively.    
    }
    \vspace{-8pt}
    \resizebox{0.99\linewidth}{!}{
        \begin{tabular}{l||c|c||c|c||c|c||c|c}
        \thickhline 
         Domain & \multicolumn{2}{c||}{Training Labels} &\multicolumn{2}{c||}{Source (Sharp)} & \multicolumn{2}{c||}{Target ( Blur)} & \multicolumn{2}{c}{Average} \\
         \hline
         Method & Sharp & Blur & mAP@0.5:0.95 & mAR@0.5:0.95 &  mAP@0.5:0.95 & mAR@0.5:0.95 &  mAP@0.5:0.95 & mAR@0.5:0.95 \\
        \thickhline 
        \hline
        Base (I)~\cite{geng2021dekr}& & \checkmark & 48.2 & 55.8 & 36.1 & 47.1 &  42.2& 51.4 \\
        Base (E)~\cite{geng2021dekr}& & \checkmark & 34.5 & 45.6 & 34.5 & 45.6 & 34.5  & 45.6 \\
        Base (I+E)~\cite{cho2025benchmark} &  & \checkmark & 55.1 &  60.6 & 54.2 & 59.8 & 54.6 & 60.2 \\
        \hline
        Base (I)~\cite{geng2021dekr} & \checkmark & \checkmark & 68.9  & 73.4  & 53.1 & 61.4 &  61.0 & 67.4 \\
        Base (E)~\cite{geng2021dekr} & \checkmark & \checkmark & 45.8 & 56.1 & 45.8 & 56.1 & 45.8 & 56.1\\
        Base (I+E)~\cite{cho2025benchmark} & \checkmark & \checkmark & 74.5 & 77.9 & 58.8 & 65.1 & 66.6 & 71.5 \\  
        \hline
        \hline
                Base (I)~\cite{geng2021dekr} & \checkmark & & 62.5 & 67.2 & 28.6 & 32.7 
 & 45.6 & 50.0 \\
        Base (E)~\cite{geng2021dekr} &\checkmark & & 43.5 & 49.7 & 43.5 & 49.7 & 43.5 & 49.7 \\
        Base (I+E)~\cite{cho2025benchmark} & \checkmark & &   \textbf{67.9} & \textbf{72.0} & 37.5 & 40.5 & 
        \underline{52.7} & \underline{56.2} \\
        \hline
        DualTeacher (I)~\cite{ai2024domain} & \checkmark & & 45.9 & 52.6 &24.4 & 29.8 & 35.2 & 41.2 \\
        DualTeacher (E)~\cite{ai2024domain} & \checkmark & & 33.5 & 41.6 & 33.5 & 41.6 & 33.5 &41.6 \\
        DualTeacher (I+E)~\cite{ai2024domain} &  \checkmark & & 60.1 & 64.7  & 34.6 & 37.5 & 47.4 & 51.1 \\
        UDA-HE (I)~\cite{kim2022unified} &\checkmark & & 46.6 & 53.4 & 18.0 & 23.3 & 32.3  & 38.4 \\
        UDA-HE (E)~\cite{kim2022unified} & \checkmark & & 29.8 & 38.5 &  29.8 & 38.5 & 29.8&  38.5  \\
        UDA-HE (I+E)~\cite{kim2022unified} & \checkmark & & 61.8 & 66.6 & 36.4 & 40.6 & 49.1 & 53.6 \\
        \hline
        \textbf{Ours-Teacher} & \checkmark & & - & - & \textbf{52.9} & \textbf{58.2} & - & -\\
        \textbf{Ours} & \checkmark & & \underline{64.2} & \underline{68.1} & \underline{51.6} & \underline{57.5} & \textbf{57.9} & \textbf{62.8}\\
        \thickhline
        \end{tabular}
             }
    \label{tab:multi_pose}
    \vspace{-10pt}
\end{table*}

These two sets of poses are combined and passed through non-maximum suppression (NMS) to generate the final poses. For these final poses, both  heatmap mask and offset mask are calculated based on the confidence from both networks. If either $C'$ or $C$ is lower than mutual masking threshold, $th'$, we mask out that region.
\begin{equation}
    \begin{aligned}
&M'_{k}(x, y) ^{\mathcal{H}} = \prod_{i \in d_k(x,y)^{\mathcal{T}}} m^{\mathcal{T}} \times \prod_{j \in d_k(x,y)^{\mathcal{S}}} 
m^{\mathcal{S}} \\ 
m^{\mathcal{T}} = &
\begin{cases} 
  1, &  min(C(P_i^{\mathcal{T}}), C'(P_i^{\mathcal{T}})) \geq th'\\ 
  0, & else 
\end{cases} \\
m^{\mathcal{S}} = &
\begin{cases}
  1, & min(C(P_i^{\mathcal{S}}), C'(P_i^{\mathcal{S}})) \geq 
  th' \\
  0, & else
\end{cases}
 \end{aligned}
\label{equ: two_teacher_masking}
\end{equation}

We can calculate \( M'_{k}(x, y) ^{\mathcal{O}} \) in the same way as \( M'_{k}(x, y) ^{\mathcal{H}} \), as done in the previous section (Sec.~\ref{sec:stage2}).
The mutual mask $M'$ is applied to the final poses to exclude unreliable predictions when calculating the loss, $\mathcal{L}_s$ between pseudo label and prediction, similar to (Eqs.~(\ref{equ:sup_loss}),~(\ref{equ: masked_heatmap_loss}), and~(\ref{equ: masked_offset_loss})) in the previous stage. This masking strategy enhances the performance of the teacher network during adaptation to the target domain, improving its ability to handle real blur in the target domain.

\noindent
\textbf{Student Network Training.}
We now proceed to retrain the student network using the teacher network that has been improved and well-adapted to the target domain. As before, we use the mutual uncertainty masking (Eqs.~(\ref{equ:confidence_crosscheck}) and~(\ref{equ: two_teacher_masking})), but this time, instead of training the teacher network with pseudo label, we focus on training the student network. This ensures that the student network benefits from the improved adaptation of the teacher network to the target domain while considering the uncertainties introduced by both networks. 
\section{Experiments}
\label{sec:experiments}

\subsection{Dataset}

\textbf{EHPT-XC Dataset}. The EHPT-XC~\cite{cho2025benchmark} dataset consists of 158 diverse sequences, each containing pixel-wise aligned and temporally synchronized event streams, sharp images and blurry images captured using the triplet camera system.  
The dataset includes 38K 2D keypoints, with 14 keypoints per person, along with bounding boxes and track IDs.
For the unsupervised domain adaptation setting, we conceptually divide EHPT-XC into three subsets: train-source, train-target, and test. The train-source subset consists of 47 sequences, each containing 100 sharp image-event stream pairs with annotations. The train-target subset contains the images and events data, but with blurry, unlabeled images. The test subset comprises 26 sequences, each providing synchronized sharp images, blurry images, and event streams.

\subsection{Implementation Details}
The input image size is set to 512 $\times$ 512. EHPT-XC dataset training is conducted on 2 TITAN RTX GPUs and batch size is 5. The training times for Stages 1–4 are 8 hours, 12 hours, 12 hours, and 12 hours, respectively.
The learning rate is 1e-3 during total 100 epochs. 
For blur augmentation, we used $N_E$=5.
In addition to motion-aware augmentation, we applied both sharp and blurry images with random rotation(-30 $\sim$ 30), horizontal flip with 0.5 probability, random scale (0.75 $\sim$ 1.5), random translation(-40 $\sim$
40 pixels) in both x, y directions. During pose estimation, we select top 30 center point locations from center heatmap and filter out center of which heat value is smaller than the keypoint threshold, 0.03 in this experiments. We set keypoints per human, $K= 14$.
$near(p)$ is set to correspond to a square region with a side length of 8 around the 2D location $p$.

\begin{figure*}[t!]
\begin{center}
\includegraphics[width=.9\linewidth]{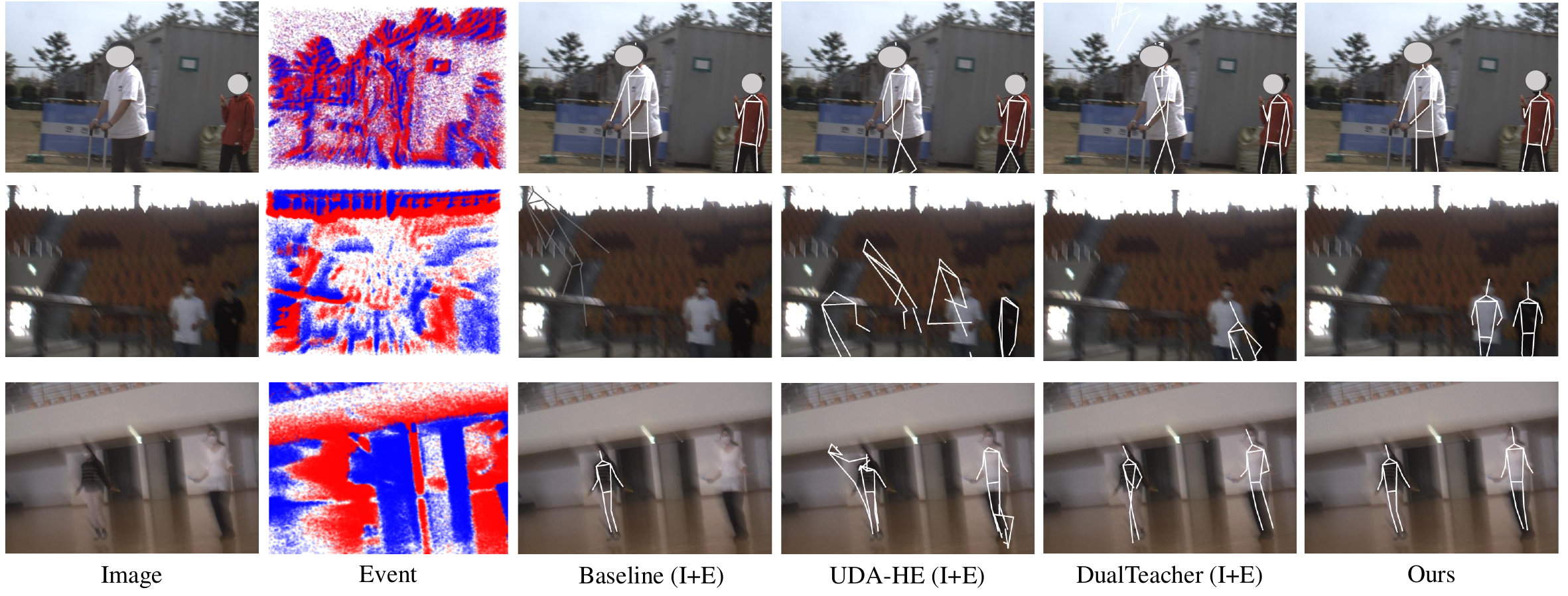}
\vspace{-9pt}
\caption{Comparison of UDA-HE~\cite{kim2022unified}, DualTeacher~\cite{ai2024domain}, baselines, and our method. The first row represents a scene with minimal blur, while the second and third rows depict scenes with severe blur.}
\label{fig:comparison_with_others}
\end{center}
\vspace{-17pt}
\end{figure*}

\subsection{Experimental Results}
We establish a baseline by using the same network architecture as our model while varying only the input modality. For some baseline settings, training was conducted using both sharp images and target-domain labels, which we refer to as the oracle setting. Additionally, to compare our approach with existing domain-adaptive human pose estimation methods, we evaluated DualTeacher~\cite{ai2024domain} and UDA-HE~\cite{kim2022unified}. Since these are image-based methods, we extend the experiments by incorporating various modalities. 
Since DualTeacher~\cite{ai2024domain}'s original augmentation is tailored for low-light conditions, we substituted it with standard augmentations.

\noindent
\textbf{Quantitative Results.}
Table~\ref{tab:multi_pose} presents a quantitative comparison between the baselines, other methods, and our approach.
Among the baselines, image-based estimation suffers the largest accuracy drop, while event-based estimation is more resilient. The event modality has the smallest domain gap due to its high-resolution motion information, but it underperforms in the sharp domain.
Domain-adaptive methods~\cite{kim2022unified, ai2024domain} show improved performance over baselines in both multi-modal and image-based settings. However, since these methods were not originally designed for the blurred domain, the significant data distribution gap leads to sub-optimal performance. In contrast, our approach effectively bridges the domain gap early on through augmentation and a multi-modal teacher network, leveraging pseudo-label refinement to achieve superior results. Notably, our method achieves performance comparable to the multi-modal oracle.

\noindent
\textbf{Qualitative Results.}
Figure~\ref{fig:comparison_with_others} compares our method against multi-modal baselines and domain-adaptive methods. Baselines without domain adaptation experience significant performance degradation as blur intensifies. While domain-adaptive methods achieve some level of pose estimation, they produce a high number of false positives. In contrast, our approach demonstrates superior performance regardless of the blur intensity, as qualitatively observed.

\begin{table}[t]
\begin{center}
\caption{Ablation study of the proposed method. MBA denotes the motion-aware event-based blur augmentation. Since Stage 3 trains only the teacher network, we report the results of performing both Stage 3 and Stage 4 together.}
\vspace{-8pt}
\resizebox{.99\linewidth}{!}
{
\renewcommand{\tabcolsep}{10.0pt}
\begin{tabular}{l|ccc|cc}
\hline
& \multirow{2}{*}{MBA} & \multirow{2}{*}{Stage 2} & \multirow{2}{*}{Stage 3 \& 4} & \multicolumn{2}{c}{Target (Blur)} \\
& & & & mAP & mAR \\
\hline
(A) &  &   &  & 37.5 & 40.5\\
(B) & \darkgreencheck & &  & 40.4 & 46.5  \\
(C) & & \darkgreencheck &  &  49.1 & 54.3 \\
(D) & \darkgreencheck  & \darkgreencheck & & \underline{50.4} & \underline{55.5} \\
(E) & \darkgreencheck & \darkgreencheck  & \darkgreencheck & \textbf{51.6} & \textbf{57.5}  \\
\hline
\end{tabular}
}
\label{tab:able_components}
\end{center}
\vspace{-27pt}
\end{table}

\begin{figure}[t]
    \centering
    \includegraphics[width=0.93\linewidth]{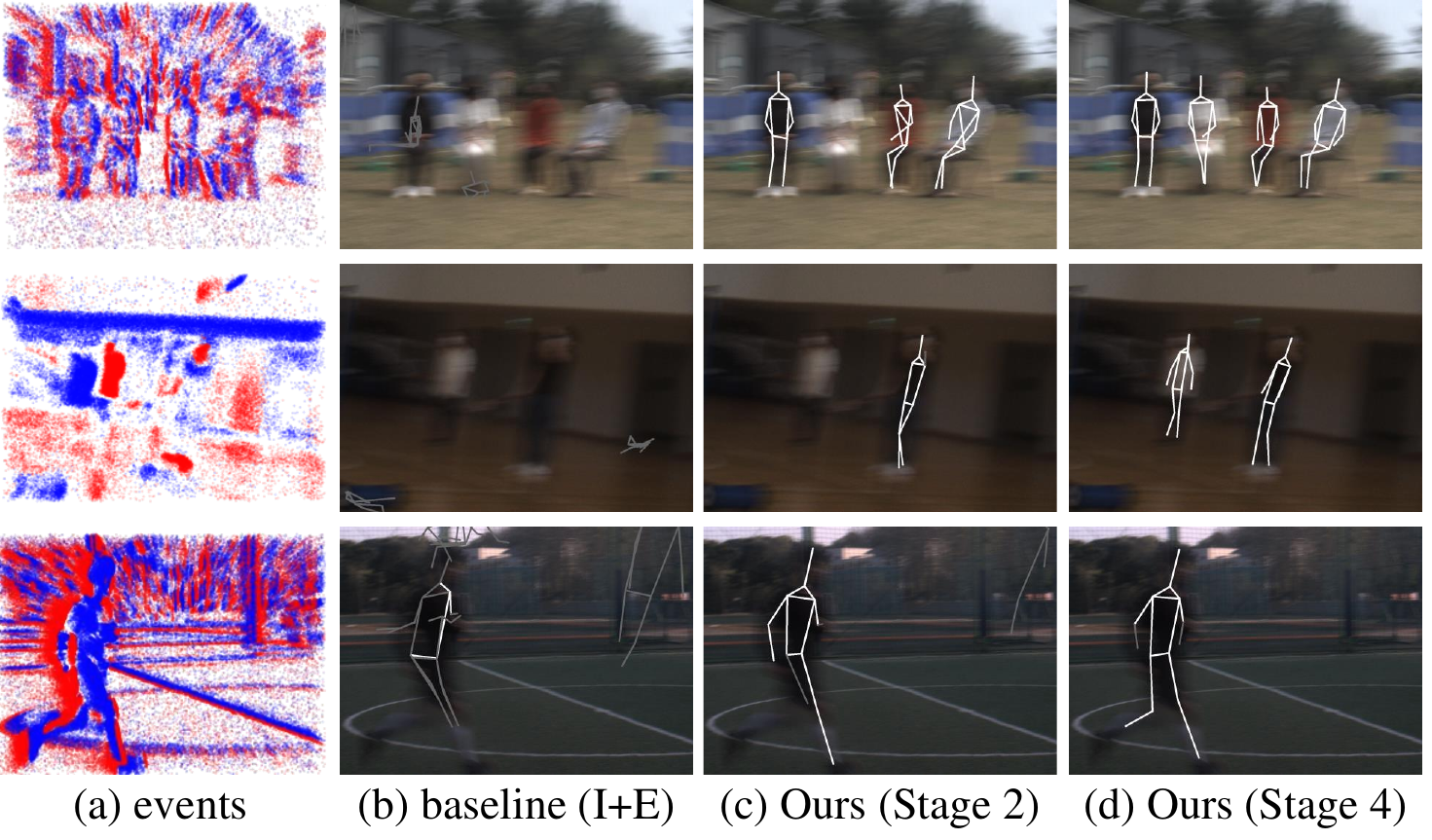}
    \vspace{-7pt}
    \caption{Qualitative comparison involving performance of each stage of our methods with baseline (image + event), stage 2 and stage 4 model.  Ours refers to the student network.}
    \label{fig:qual}
    \vspace{-3pt}
\end{figure}

\begin{table}[t]
\begin{center}
\caption{Effectiveness of motion-aware event-based blur augmentation. We report the result of the teacher network from Stage 1. Note that teacher network is trained on only source domain.}
\vspace{-4pt}
\resizebox{.99\linewidth}{!}
{
\renewcommand{\tabcolsep}{13pt}
\begin{tabular}{l|cc}
\hline
\multirow{2}{*}{Method} & \multicolumn{2}{c}{Target (Blur)} \\
& mAP & mAR \\
\hline
Baseline &  46.5 & 50.4 \\
+ Motion Aug.  & \underline{46.7}  & \underline{51.9}\\
+ Motion Aug. + Blur Translation  & \textbf{46.9} & \textbf{52.2} \\
\hline
\end{tabular}
    }
\label{tab:able_aug}
\end{center}
\vspace{-24pt}
\end{table}

\section{Ablation Study and Analysis}
\label{sec:ablation}

\textbf{Effectiveness of the modules.} Table~\ref{tab:able_components} shows the 
ablation study results for different components, evaluating adaptation performance in the target domain as each proposed module is progressively added. The results confirm that each module and training strategy contributes to overall performance improvement. Notably, combining augmentation with pseudo-label-based learning significantly enhances accuracy, while additional refinement in Stages 3 and 4 further improves final performance. 
The improvement in target domain adaptation through pseudo-label refinement across stages is more clearly illustrated in Fig.~\ref{fig:qual}.

\noindent
\textbf{Effectiveness of the motion-aware augmentations} As shown in Table~\ref{tab:able_aug}, compared with baseline, motion augmentation improves mAP by 0.4, and when combined with blur translation, it achieves a total improvement of 1.8.

\begin{table}[t]
\begin{center}
\caption{Analysis of the teacher network components. The results of the teacher network \( \mathcal{T} \) in Stage 1 are presented.}
\label{tab:able_mask}
\vspace{-8pt}
\resizebox{.94\linewidth}{!}
{
\renewcommand{\tabcolsep}{13.5pt}
\begin{tabular}{c|l|cc}
\hline
& \multirow{2}{*}{Modality} & \multicolumn{2}{c}{Target (Blur)} \\
& & mAP & mAR \\
\hline
\multirow{3}{*}{w/o. $\mathcal{T}_R$} & $\mathcal{T}_I$ &   30.5 & 37.2 \\
& $\mathcal{T}_E$ &   \underline{43.5} & \underline{49.7} \\
& $\mathcal{T}_{I+E}$ &  40.4 & 46.5 \\
\hline
\multirow{4}{*}{w. $\mathcal{T}_R$} & $\mathcal{T}_I$ &   32.7 & 37.0 \\
& $\mathcal{T}_E$ & 43.1 & 48.8 \\
& $\mathcal{T}_{I+E}$ &  39.6 & 46.2 \\
& $\mathcal{T}_{I+E}$ + $\mathcal{T}_E$ (Ours)&  \textbf{46.9} & \textbf{52.2} \\
\hline
\end{tabular}
}
\label{tab:teacher_components}
\end{center}
\vspace{-15pt}
\end{table}

\begin{table}[t]
\begin{center}
\caption{Analysis of the method for merging the outputs of the sub-teacher networks in the teacher network. All methods use \( \mathcal{T}_{I+E} \) and \( \mathcal{T}_E \) as sub-teachers. The results of the teacher network \( \mathcal{T} \) in Stage 1 are presented.}
\vspace{-8pt}
\resizebox{.82\linewidth}{!}
{
\renewcommand{\tabcolsep}{16.5pt}
\begin{tabular}{l|cc}
\hline
\multirow{2}{*}{Method} & \multicolumn{2}{c}{Target (Blur)} \\
& mAP & mAR \\
\hline
Mean Heat Map &  \underline{45.4} & \underline{50.8} \\
Spatial Attention~\cite{woo2018cbam}  & 41.9 & 48.3\\
Refinement Network  & \textbf{46.9} & \textbf{52.2} \\
\hline
\end{tabular}
}
\label{tab:able_refine_with_fusion}
\end{center}
\vspace{-15pt}
\end{table}

\noindent
\textbf{Analysis of teacher network components.}
The experimental results on the components of the teacher network are presented in Table~\ref{tab:teacher_components}. The results indicate that relying on a single sub-teacher component fails to fully exploit the refinement module, as each modality exhibits strengths and weaknesses depending on the scene conditions.
Our teacher network consists of a multi-modal (I+E) sub-teacher, an event (E) sub-teacher, and a refinement module, allowing it to leverage the benefits of multi-modal learning while effectively handling noisy labels. As a result, it achieves the highest performance.

\noindent
\textbf{Analysis of the refinement network.}
To refine the final output based on the results of the sub-teacher networks, we design and incorporate a refinement network. As alternatives, one could average the heatmaps from both networks or apply spatial attention~\cite{woo2018cbam} mechanisms. However, as shown in Table~\ref{tab:able_refine_with_fusion}, our method, which computes high-representation attention across channels for fusion, achieves the best performance among all approaches.

\begin{table}[t]
    \caption{Hyper-parameter analysis of the masking threshold. \( th \) and \( th' \) denote the thresholds for Stage 2 and Stage 4, respectively.}
    \vspace{-7pt}
    \setlength\tabcolsep{11.2pt}
    \begin{minipage}[b]{.47\linewidth}
        \centering
        \resizebox{0.99\linewidth}{!}{
            \begin{tabular}{ccc}
            \hline
            \rowcolor[rgb]{0.95,0.95,0.95}
            \multicolumn{3}{c}{Stage 2} \\ 
            \hline
            $th$   & mAP & mAR \\ 
            \hline
            0.05 & 47.2 & \underline{52.8} \\
            0.1 &  \textbf{50.4} & \textbf{55.5} \\
            0.2 & \underline{47.7} & 52.5 \\
            0.3 & 45.7 & 51.0\\
            \hline
            \end{tabular}
        }
    \end{minipage}
    \hfill
    \begin{minipage}[b]{.47\linewidth}
        \centering
        \resizebox{0.99\linewidth}{!}{
            \begin{tabular}{ccc}
            \hline
            \rowcolor[rgb]{0.95,0.95,0.95}
            \multicolumn{3}{c}{Stage 4} \\ 
            \hline
            $th'$   & mAP & mAR \\ 
            \hline
            0.05 & 49.7 & 56.3 \\
            0.1 &  \textbf{51.6} & \textbf{57.5}\\
            0.2 & 49.7 & \underline{56.9} \\
            0.3 & \underline{50.3} & \underline{56.9}\\
            \hline
            \end{tabular}
        }
    \end{minipage}
    \vspace{-9pt}
    \label{tab:able_thres}
\end{table}

\noindent
\textbf{Threshold Analysis.}
Table~\ref{tab:able_thres} presents the performance variations based on the threshold for training the student network in Stages 2 and 4. Notably, when relying on a single-teacher network, the model becomes more sensitive to threshold selection, leading to performance fluctuations. In contrast, with a dual-network framework, the filtering process produces more reliable pseudo-labels, resulting in greater robustness across a wider range of thresholds.

\begin{figure}[t!]
\begin{center}
\includegraphics[width=.92\linewidth]{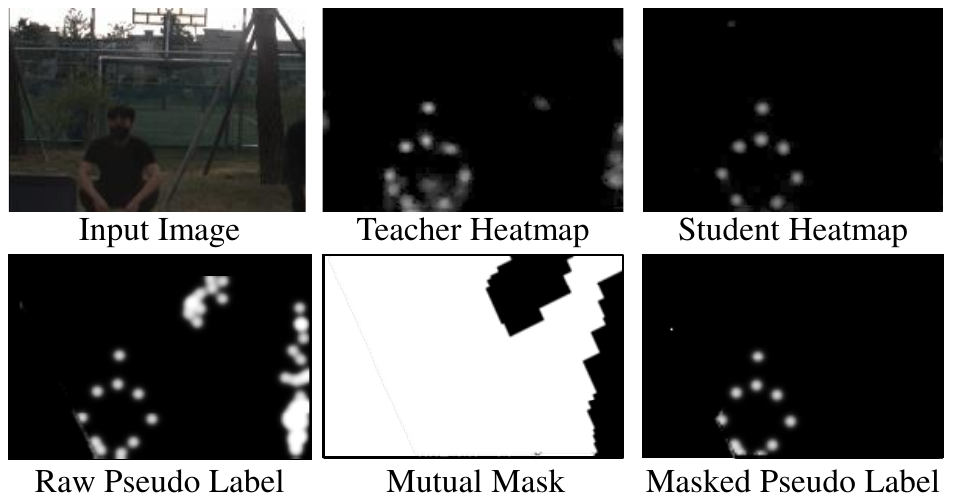}
\vspace{-6pt}
\caption{Pseudo label with mutual uncertainty masking.}
\label{fig:heat_map_mask}
\end{center}
\vspace{-19pt}
\end{figure}

\noindent
\textbf{Effectiveness of mutual uncertainty masking.} Figure~\ref{fig:heat_map_mask} illustrates the effectiveness of mutual uncertainty masking. In many cases, either the teacher or the student network predicts the correct answer while the other produces an incorrect one. Pseudo-labeling based on a single network is vulnerable to such inconsistencies. However, our approach can extract reliable masking even from the student heatmap, effectively filtering out uncertain regions in the teacher heatmap, which has a higher capacity. As a result, the masked pseudo-labels become more accurate, eliminating uncertain areas and improving overall reliability.

\noindent
\textbf{Analysis of using only image data at test time.}
To apply pose estimation at real-world, aligned image-event data can be hard to acquired. So we investigated the performance using only image data at test time. We trained a final student network using only image data with the pseudo labels acquired from the multi-modal teacher network. As shown in Table ~\ref{tab:only image ours}, without event data at test time, our framework shows that the proposed method can achieve better performance from the image-only baseline.

\begin{table}[!t]
\caption{Performance of image-only student trained with an image+event teacher network at Stage 4.}
    \setlength\tabcolsep{4.0pt}
    \centering
    \vspace{-7pt}
    \resizebox{0.99\linewidth}{!}{
        \begin{tabular}{l||c|c||c|c||c|c||c|c}
        \hline 
         Domain & \multicolumn{2}{c||}{Training Labels} &\multicolumn{2}{c||}{Source (Sharp)} & \multicolumn{2}{c||}{Target ( Blur)} & \multicolumn{2}{c}{Average} \\
         \hline
         Method & Sharp  &  Blur & mAP & mAR &  mAP & mAR &  mAP & mAR \\
        \hline 
        Base (I) & & \checkmark & 48.2& 55.8  & 36.1 & 47.1 & 42.2 & 51.4 \\
        \thickhline
        Base (I) & \checkmark & & 62.5 & 67.2& 28.6 & 32.7 & 45.6 & 50.0\\
        \textbf{Ours (I)} & \checkmark & & \textbf{63.2} & \textbf{68.1} & \textbf{46.0 }&\textbf{52.7} & \textbf{54.6} & \textbf{60.4} \\
        \hline
        \end{tabular}
        }
    \label{tab:only image ours}
    \vspace{-9pt}
\end{table}

\section{Conclusion}
\label{sec:conclusion}

In this paper, we introduce event cameras for the first time to facilitate domain-adaptive human pose estimation from a sharp domain to a motion-blurred domain. Without relying on target-domain annotations, we propose an effective adaptation strategy leveraging motion-aware event-based blur augmentation and dual network-based mutual uncertainty masking. Our approach successfully bridges the domain gap, achieving performance comparable to that of an oracle model trained with target-domain annotations. Our work highlights the potential of event cameras for robust human pose estimation and their applicability in dynamic environments.

\section{Acknowledgments.}

This work was supported by the National Research Foundation of Korea(NRF) grant funded by the Korea government(MSIT) (NRF2022R1A2B5B03002636), and by the Institute of Information \& communications Technology Planning \& Evaluation (IITP) grant funded by the Korea government(MSIT) (No. RS-2024-00457882, AI Research Hub Project).

{
    \small
    \bibliographystyle{ieeenat_fullname}
    \bibliography{main}
}

\end{document}